\documentclass[letterpaper, 10pt, conference]{ieeeconf}  

\IEEEoverridecommandlockouts                              

\usepackage{graphicx}
\usepackage{lipsum}
\usepackage{amsmath}
\usepackage{amssymb}
\usepackage{cite}
\usepackage{algorithm}
\usepackage{svg}
\usepackage{url}
\usepackage{multirow}

\usepackage{flushend}
\usepackage[normalem]{ulem}
\usepackage[margin=0.75in]{geometry}

\usepackage{algpseudocode}

\title{\LARGE \bf
Real-Time Reinforcement Learning for Dynamic Tasks \\ with a Parallel Soft Robot
}

\author{James Avtges$^{1}$, Jake Ketchum$^{1}$, Millicent Schlafly$^{1}$, Helena Young$^{1}$ \\ Taekyoung Kim$^{2}$, Allison Pinosky$^{1}$, Ryan L. Truby$^{1,2}$, Todd D. Murphey$^{1}$ 
\thanks{$^{1}$Department of Mechanical Engineering, Northwestern University, Evanston, IL, USA.
        }%
\thanks{$^{2}$Department of Materials Science and Engineering, Northwestern University,
        Evanston, IL, USA.
        }%
\thanks{This work is supported by the US Army Research Office grant no. W911NF-22-1-0286, US Office of Naval Research grant no. N00014-21-1-2706, and the National Science Foundation under Award NSF EEC 2330040\color{black}. J.A. is supported by a National Defense Science and Engineering Graduate Fellowship. T.K. and R.L.T. acknowledge support from US Office of Naval Research grant no. N00014-22-1-2447 and Leslie and Mac McQuown through Northwestern University’s Center for Engineering Sustainability and Resilience.}
}

\begin{document}

\maketitle
\thispagestyle{empty}
\pagestyle{empty}

\begin{abstract}

Closed-loop control remains an open challenge in soft robotics. The nonlinear responses of soft actuators under dynamic loading conditions limit the use of analytic models for soft robot control. Traditional methods of controlling soft robots underutilize their configuration spaces to avoid nonlinearity, hysteresis, large deformations, and the risk of actuator damage. Furthermore, episodic data-driven control approaches such as reinforcement learning (RL) are traditionally limited by sample efficiency and inconsistency across initializations. In this work, we demonstrate RL for reliably learning control policies for dynamic balancing tasks in real-time single-shot hardware deployments. We use a deformable Stewart platform constructed using parallel, 3D-printed soft actuators based on motorized handed shearing auxetic (HSA) structures. By introducing a curriculum learning approach based on expanding neighborhoods of a known equilibrium, we achieve reliable single-deployment balancing at arbitrary coordinates. In addition to benchmarking the performance of model-based and model-free methods, we demonstrate that in a single deployment, Maximum Diffusion RL is capable of learning dynamic balancing after half of the actuators are effectively disabled, by inducing buckling and by breaking actuators with bolt cutters. Training occurs with no prior data, in as fast as 15 minutes, with performance nearly identical to the fully-intact platform. Single-shot learning on hardware facilitates soft robotic systems reliably learning in the real world and will enable more diverse and capable soft robots.
\end{abstract}

\section{INTRODUCTION}

Soft robots offer the potential for improved adaptability and safety compared to their rigid counterparts due to compliance and material redundancies \cite{rus2015design}. However, soft-actuated robots are traditionally difficult to control due to their nonlinear dynamics and high degrees of freedom \cite{polygerinos2017soft, bruder2019modeling}. Soft-actuated robot dynamics---which may be impractical or impossible to model classically---are often stochastic and vary with a number of factors including strain, actuator fatigue, and even manufacturing processes \cite{mbcSoftRobtRus, Thuruthel2019}.

Data-driven methods such as reinforcement learning (RL) have been used to circumvent many of the modeling difficulties associated with controlling soft robots \cite{Kim2021}. Prior applications of RL in this domain have explored various approximations of soft actuator dynamics, including piecewise constant curvature models, Cosserat rod models, and rigid $N$-link pendulum approximations \cite{jitosho2023reinforcement, Centurelli2022, Morimoto2021, Thuruthel2019, alessi2024pushing}. However, lower-dimensional approximations may oversimplify the dynamics of highly nonlinear actuators, which may be compounded by external forces and changes to the dynamics during use. Additionally, models parameterized using these approximations are almost exclusively used to train control policies in simulation, requiring dedicated approaches to address the sim-to-real gap. Existing approaches also often assume quasi-static behavior and focus on relatively simple tasks such as reaching or tracing with continuum arms.

When learning approaches are applied to soft robot control, current works often constrain the size or dimensionality of the configuration space to regions with smaller deformations, or where their dynamics are approximately linear \cite{Ketchum2023, jitosho2023reinforcement, Morimoto2021}. This may include directly constraining actuator outputs, choices of robot orientation, and evaluating control approaches only on undamaged, unloaded actuators. \color{black} While these techniques are effective, utilizing the full variety of soft robots' diverse configuration spaces despite their stochastic and nonlinear dynamics that may be impossible to simulate is paramount for producing behaviors that involve buckling, large deformations, or actuator breakages. Exploiting these unique properties of soft robots has the potential to enable more diverse task learning, biomimetic behaviors, and adaptation to loading conditions or damage.

\color{black}
Our experimental platform is a six degree-of-freedom (DoF), soft-actuated parallel mechanism with a structure similar to that of a Stewart platform \cite{stewart}. The struts of the platform are motorized soft actuators based on 3D-printed Handed Shearing Auxetics (HSAs) that lengthen and shorten upon the turn of a servo motor \cite{HSAScience, Kim2024}. In this work, we balance a sliding puck on the platform, at both the center and at arbitrary coordinates. We utilize deep RL to train closed-loop controllers and benchmark multiple model-based and model-free RL frameworks. Our training occurs fully on-hardware without simulation or bootstrapped data.

\color{black}

Learning in real-time on hardware has its challenges \cite{dulac2019challenges, ibarz2021train}, some unique to soft robotics. The limited lifespan of flexible, strain-dependent actuators makes sample efficiency and adaptability valuable attributes \color{black} for minimizing and adapting to actuator damage throughout training \cite{Morimoto2021}. Our approach combines a classical kinematics model for a rigid Stewart platform with a deep RL controller to efficiently learn nonlinear control policies for the system. We train all of our policies using \textit{single-shot reinforcement learning}, a non-episodic problem formulation for continual learning without any environment resets. Learning in a singular real-time episode may be necessary for a variety of tasks well-suited to soft robotics---such as functioning in extreme environments or human interactions---where leveraging prior data, simulations, and reset routines may be undesirable or impossible. While episodic soft robot RL has been implemented on-hardware in works such as \cite{Morimoto2021, hardwareinloop, raffin2022smooth}, and reset-free hardware learning has been implemented in works such as \cite{wu2023daydreamer, allie2025noodlebot}, to the best of our knowledge this work is the first to accomplish all of the above in a soft-actuated robot.\color{black}

\begin{figure}[hbtp]
    \centering
    \includegraphics[width=.85\linewidth,keepaspectratio]{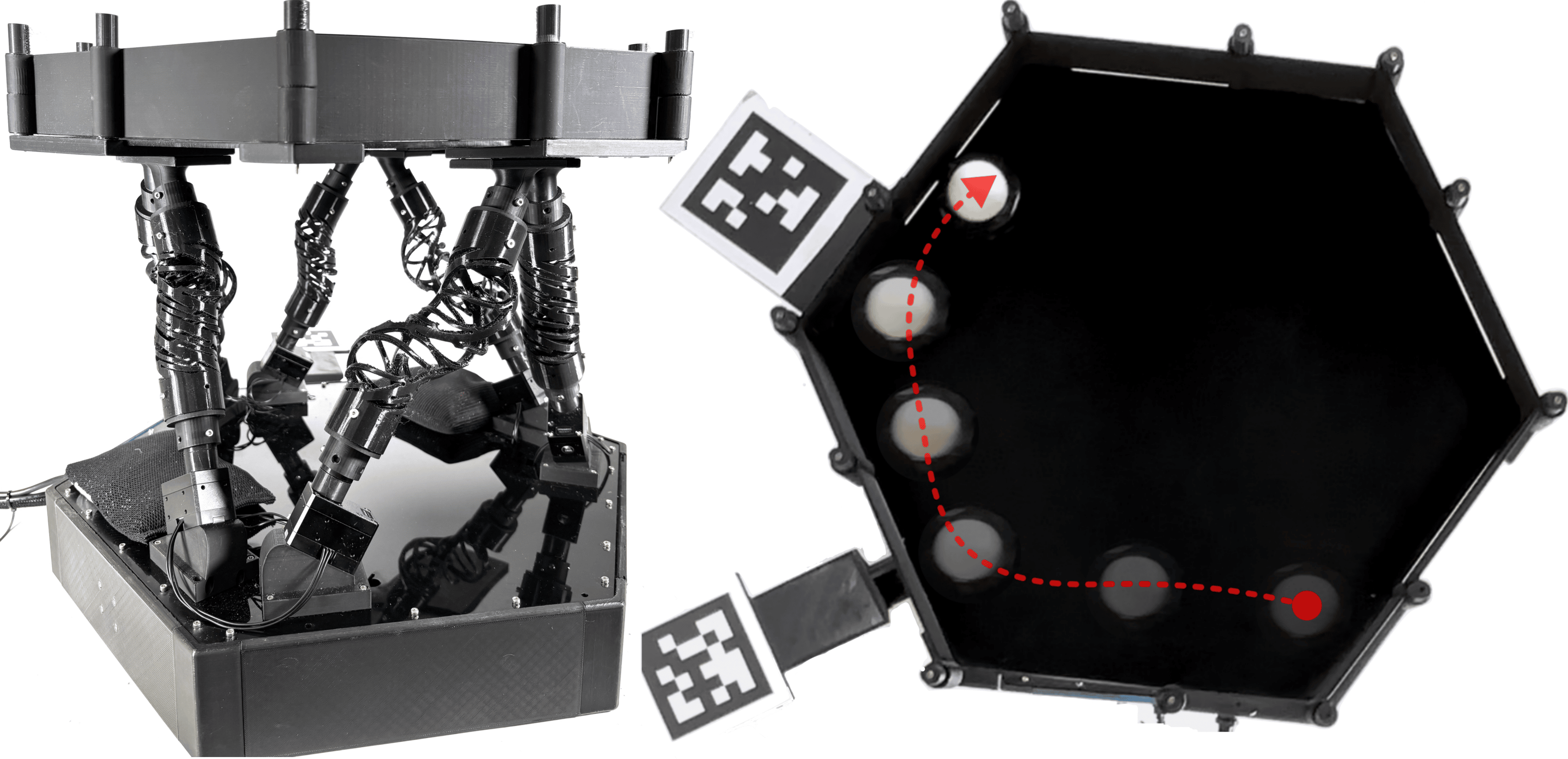}
    \caption{\textbf{Learning dynamic tasks in a parallel soft robot.} The experimental platform: a 6-DoF parallel soft robot. A puck is placed on the platform; the learning objective is to balance the puck at arbitrary equilibria. Dynamic balancing is achieved despite the nonlinear behavior of the soft actuators, including buckling and breaking the HSAs.}
    \label{fig:photo}
\end{figure}

An additional challenge with learning a balancing task in single-shot episodes is that reliably learning control policies is not necessarily an inevitable outcome in reasonable timescales---catastrophic failure modes do exist. While the balancing puck is constrained from falling off the platform, without access to a reset routine, either learned or programmed, the puck can become stuck in a corner during training, providing little to no variance in data required for learning. This is especially \color{black} relevant when learning to balance at arbitrary coordinates on the platform, where the reward landscape is parameterized by an observable setpoint that may not be near the platform's center. Other works conducting soft-actuated RL such as \cite{Morimoto2021, hardwareinloop, Centurelli2022, raffin2022smooth} structure their experiments such that providing zero control results in an intervention-free environment reset---such as orienting a continuum arm downwards---stable states such as these do not exist in our experimental platform with an unarticulated sliding puck. \color{black}

To overcome this challenge, we employ a curriculum learning approach based on expanding neighborhoods of a known equilibrium: the platform center. With this we reliably learn to balance in a single deployment, without it the task is impossible to accomplish consistently, or at all.

Furthermore, we show how RL can learn to balance the puck despite major changes to the robot's dynamics during training. We buckle half of the HSAs, introducing singularities, hysteresis, and an out-of-distribution configuration. We also damage the platform by cutting through half of the HSAs with bolt cutters. Despite these alterations, RL attains evaluation performance indistinguishable to the default case.

The contributions of this work are:

\begin{itemize}
    \item Demonstrations and benchmarking of single-shot learning for dynamic tasks on a soft robot,
    \item Introduction of a curriculum learning procedure to improve single-shot learning outcomes, and
    \item An illustration of a setting where RL can accommodate changing dynamics such as buckling or breaking actuators during single-shot training.
\end{itemize}

\color{black}
\section{RELATED WORK}

\subsection{Reinforcement Learning and Control in Soft Robotics}

RL has been applied in several works to learn data-driven control for soft robotics \cite{Centurelli2022, Thuruthel2019, Morimoto2021, jitosho2023reinforcement, Kim2021}. While a variety of techniques have been explored in both model-based and model-free RL, the majority of these works consider quasi-static tasks such as reaching to target points or trajectory tracking for continuum arms \cite{Centurelli2022, moderncontrolmethods}. However, Jitosho et al.\ do demonstrate dynamic reaching and swinging behaviors in the real world using a pneumatic artificial muscle with near-linear dynamics \cite{jitosho2023reinforcement,linearPAM}. They learn control policies in simulation using an $N$-link rigid pendulum approximation, and overcome the sim-to-real gap using domain randomization and by modeling actuator dynamics \cite{jitosho2023reinforcement}. Fischer et al.\ demonstrate impressive throwing and pick-and-place movements using a continuum arm; although they use a proportional-derivative controller for their control law \cite{dynamicTaskSpaceFisher}. Centurelli et al.\ apply model-free, on-policy RL for dynamic trajectory tracking with variable payloads, trained also in simulation with an approximate dynamical model using Recurrent Neural Networks \cite{Centurelli2022}.

Reinforcement learning using model-free, off-policy methods was conducted in \cite{Morimoto2021} for reaching tasks. The authors do demonstrate learning in the real world alongside sim-to-real transfer; however, actuator breakages complicated the 20-hour long training process \cite{Morimoto2021}.

An example of RL applied to a two-DoF crawler is provided by \cite{hardwareinloop}, using model-free, off-policy RL to locomote. This work demonstrates learning trained on hardware over several trials, and the authors benchmark their results against random actions. Training takes up to 5 hours. In contrast, HSAs provide a convenient option for training higher-dimensional RL systems on hardware as they remain functional at high deformation rates and force outputs for longer than many other soft actuators \cite{kaarthik2022motorized}. We learn dynamic balancing in under 15 minutes and have logged over 80 total operating hours without actuator breakages. \color{black}

\subsection{Handed Shearing Auxetics}

HSAs are a class of architected materials \cite{HSAScience} that can be designed for use as electrically-driven, high-force soft robotic actuators. They have been used as soft actuators for a wide range of systems, including soft robotic grippers, soft quadrupeds, crawling continuum robots, and artificial muscles \cite{HSAScience, Truby2021, kaarthik2022motorized, Kim2024}. In the case of cylindrical HSAs, the architecture and auxetic properties of HSAs result in a coupling of extensional and rotational motion. When one end of the HSA is rotationally constrained, an applied torque at the opposite end drives linear motion. Thus, by supplying torques with a servo motor, HSAs become more efficient transducers compared to other soft actuation methods \cite{Evenchik2023} that are capable of linear extension at higher force output and faster actuation speeds \cite{Kim2024, kaarthik2022motorized}. Multi-DoF actuators can be constructed from assemblies of HSAs with opposite handedness \cite{HSAScience}. HSAs can be equipped with sensors for enhanced state feedback, utilizing soft capacitive sensors, fluidic sensors, and cameras \cite{zhang2022vision, truby2022fluidic}. Recent progress in 3D printing HSAs with techniques like digital projection lithography using polyurethane photoresins or fused deposition modeling with thermoplastic polyurethane (TPU) filaments \cite{Truby2021, Kim2024} have enabled HSAs to be fabricated with various mechanical properties and sizes. Like all soft material-based actuators, HSAs exhibit nonlinear viscoelastic behaviors over their operational lifespan. This results in a reduction of their stiffness over time and hysteresis in actuation strain, input torque necessary for actuation, and their force output \cite{Truby2021}. These nonlinear mechanical effects can be exacerbated by higher strain rates, increased loads, and plastic changes due to processes like material creep.

\subsection{Single-Shot and Curriculum Learning}

Single-shot learning---a problem formulation wherein skills are acquired by an agent through one continual deployment with no episodes or resets---is desirable for many real-world tasks where resetting the environment is impractical or impossible \cite{bruce2017one, Berrueta2024, montgomery2017reset, yang2019single}. In contrast, most RL frameworks repeatedly reset the environment and redeploy agents episodically with varying initial conditions \cite{ibarz2021train, smith2024grow}.

Learning a task over one attempt has also been termed `single-life' RL in \cite{chen2022you}, which leverages offline data to more quickly learn novel tasks. Several other works have approached non-episodic learning in simulated benchmarks \cite{eysenbach2017leave, lomonaco2020continual}, and on robots in the real world \cite{kim2022automating, walke2023don, gupta2021reset, sharma2023self}. Many of these works learn to reset the environment alongside a task by using a task-graph representation or explicitly learning the reverse task \cite{kim2022automating, walke2023don, gupta2021reset, sharma2023self, eysenbach2017leave}.

In hardware, applications of end-to-end RL in the real world include self-driving cars as demonstrated in \cite{hester2012rtmba}, controlling an autonomous vehicle with no prior training data using a model-based learning approach. Gupta et. al.\ \cite{gupta2021reset} introduces a method for reset-free learning entirely in the real world for manipulation tasks. Real-time gait learning from scratch has been demonstrated in \cite{smith2024grow}. Gait learning in reset-free deployments has also been approached in soft robots by \cite{Ketchum2023}, using a quadruped constructed from 16 HSAs. Like our work, \cite{Ketchum2023} achieves sample efficiency by constraining the high-dimensional search space and using motion primitives to learn gaits.

To overcome the challenge of unstable learning performance in arbitrary-point balancing, we employ curriculum learning (CL) \cite{bengio2009curriculum}, a procedure in which a sequence of tasks enables the transfer of a learned representation from easier problems to more difficult ones. In reinforcement learning settings, curricula have been applied in many works in both simulated and hardware settings as described in \cite{narvekar2020curriculum}, through manually designed, automatically sequenced, or generated curriculum steps. CL has been shown to improve generalization, convergence rate, and peak performance \cite{wang2021survey}. Our approach is similar to those of \cite{selfridge1985training} and \cite{sanger1994neural}---we define an initial range of setpoints that are near a known equilibrium and gradually make them more challenging until the entire platform is covered.

\color{black}
\section{METHODS}

\subsection{Soft Robot Design \& Experimental Setup}

The robot used in this work is a six-DoF parallel manipulator constructed with the structure of a Stewart platform. It is actuated by six HSAs, which are 3D printed with fused deposition modeling (FDM) using TPU 95A filament on a Bambu X1-Carbon printer. Each actuator is fully constrained at the top of the platform and driven at the opposite end by a DYNAMIXEL XM430-W350-T servo motor. A key difference between our experimental platform and a traditional Stewart platform is the lack of ball joints at the top and bottom of the platform struts; the HSA compliance accounts for all of the kinematic behavior of the platform, which would otherwise be fully constrained. The top surface of the platform is a hexagonal piece of acrylic, where each side is 24cm and the surface is sanded to reduce friction. The platform also has 3D printed polylactic acid (PLA) walls and edges, preventing the puck from falling off and providing intermittent mechanical contact. The same model we use for control also allows the platform to be teleoperated using a SpaceMouse Compact CAD mouse. All sensing, learning, and control is interfaced using ROS2.

Two Apriltags are used, one each at the top and bottom of the platform, to provide state information for RL. They are tracked by a Logitech BRIO webcam above the platform, which also detects the balancing puck using a Hough circles algorithm. We use ray-tracing to determine the position of the puck relative to the platform center. We calculate the state of the platform-puck system at approximately 60Hz, and provide motor commands at 15Hz. 

\subsection{Learning Dynamic Tasks}

In this work, we focus on dynamic balancing tasks: at the platform center and at arbitrary points on the platform. The balancing `puck' is a solid 38mm Delrin ball inside a plastic ring that measures 5cm in diameter and provides dampening. The 50g puck experiences static, kinetic, and rolling friction.

\subsubsection{Center Balancing}

For balancing at the center, a setpoint is fixed to the origin of the platform surface for the entire single-shot learning episode. 

\subsubsection{Arbitrary-Point Balancing}

For balancing at arbitrary locations, setpoints are uniformly sampled from within a variable radius of the center, which parameterizes the curriculum learning. We provide a small initial radius for the curriculum and the sampling radius expands to cover the whole platform during the first half of the training episode, after which it remains constant. New setpoints are generated once every 60s. Sampling setpoints closer to the center at the start of training prevents the puck from becoming stuck in corners before it learns the relationship between coordinates and rewards. Without such an approach, training is not stable enough to learn on practical timescales for this benchmark system, less than a day of continuous training.

\begin{figure}[t]
\begin{algorithm}[H]

\caption{Curriculum for Arbitrary-Point Balancing}
\label{sampling}
\begin{algorithmic}[1]
\State \textbf{Initialize} step count, platform radius $R$, min. radius $\lambda_0$, curriculum speed $\gamma$
\While{training}
\If{step count $\mathbin{mod}$ steps per setpoint $=$ 0}
    \State setpoint $\gets$ \Call{curriculumSample}{step count}
\EndIf

\EndWhile
\State
\Function{curriculumSample}{step count}
\State $ \lambda = \min(\text{step count} \cdot \gamma + \lambda_0, 1) \cdot R$ 

\State $\phi, \beta \leftarrow{U[0, 1)} $ \Comment{Uniformly sample}
\State $x= \lambda \beta \cdot \sin(2\pi\phi)$, $y= \lambda \beta \cdot \cos(2\pi\phi)$
\State \textbf{return} x, y
\EndFunction
\end{algorithmic}
\label{curriculum}
\end{algorithm}
\vspace{-30pt}
\end{figure}
\color{black}

\subsection{Geometric Model for Simplified Control}

To control the platform, we use a geometric model of a rigid Stewart platform with dimensions that approximate the physical robot. This allows us to provide approximate commands in the end-effector frame, reducing the control dimensionality from six to two. For a given roll, pitch, and constant z-axis coordinate, the lengths of each of the struts are calculated as if the platform were rigid. The inverse kinematics of the model are \[L = \lvert \lvert RP - B + T \lvert \lvert\] where $L \in \mathbb{R}^6$ represents the strut lengths to achieve the desired rotation $R \in SO(3)$ and translation $T \in \mathbb{R}^{3\times6}$. Matrices $P$ and $B \in \mathbb{R}^{3\times6}$ in each column contain the respective coordinates for each strut expressed in the top and bottom coordinate frames of the platform. Each column in $T$ contains the commanded translation of the platform relative to its base. Implementation details were provided by \cite{stewartPy}.  

The calculated strut lengths are then translated into rotational commands for the servo motors. Before installing the HSAs, we measure their extension lengths and make a linear-extension assumption. We show in the nominal length and extension rate of two representative HSAs in Table \ref{hsas}: one freshly printed and one with approximately 80h of usage.
\textbf{\begin{table}[hbtp]
\begin{center}
\caption{HSA parameters before and after training.}
\begin{tabular}{|l|l|l|}
\hline
\textbf{HSA Type} & \textbf{Nominal Length} & \textbf{Extension Rate} \\ \hline
\textbf{No Usage} & 136 mm                  & 12.3 mm/rad             \\ \hline
\textbf{Worn}     & 124 mm                  & 13.8 mm/rad             \\ \hline
\end{tabular}
\label{hsas}
\end{center}
\end{table}}
\vspace{-10pt}

There are key differences between the rigid geometric model and the soft Stewart platform used for experiments. The lack of ball joints makes the kinematic relationship inaccurate. The relevant geometric parameters for the soft Stewart platform in Table \ref{hsas} change with usage, producing further inaccuracies. Additionally, the platform oscillates when controlled at high frequencies---with actuator dynamics that are subject to change as HSAs wear, buckle, and break. Our RL approach functionally learns the residuals in the rigid model plus the nonlinearities in the HSAs and puck dynamics.

\begin{figure*}[t]
    \centering
    \includegraphics[width=0.9\textwidth,keepaspectratio]{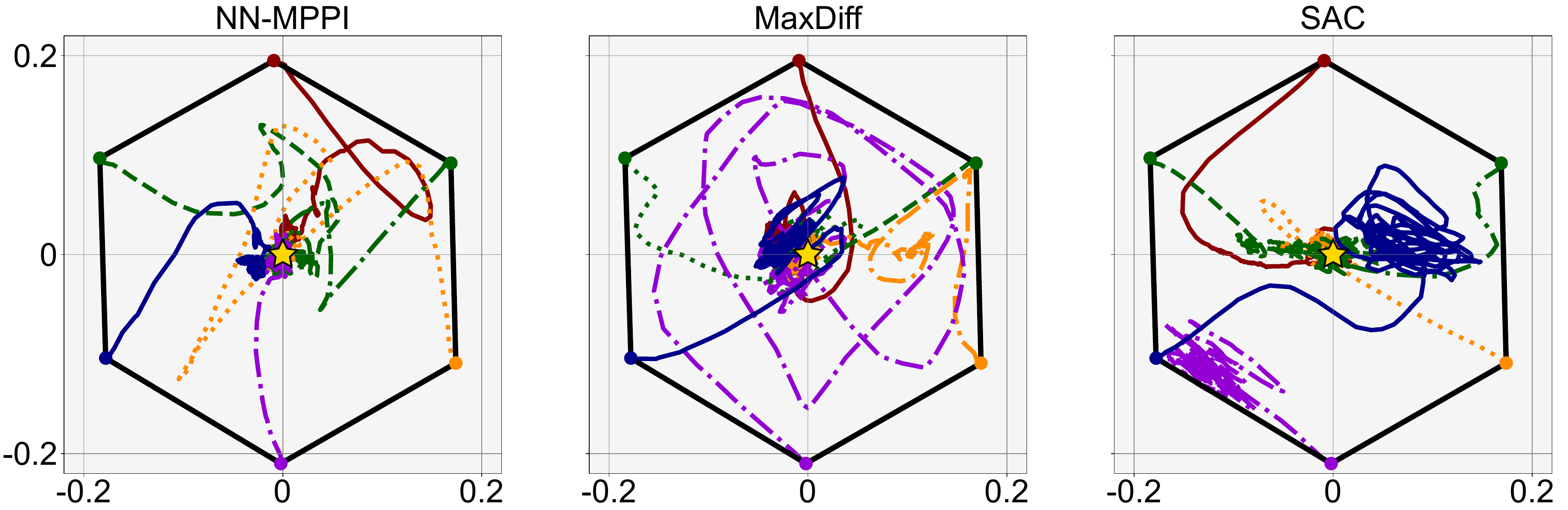}
     \caption{\textbf{Evaluation trajectories for center balancing.} We evaluate five seeds for each model with the puck starting from each corner of the platform with no initial tilt. We display representative evaluation trajectories, with each color representing a different seed. NN-MPPI and MaxDiff consistently balance at the center, while certain seeds of SAC fail to achieve the task entirely. Without explicitly supplying path-dependent rewards, the sample paths to balance can be circuitous and nonlinear.}
    \label{fig:center_eval}
    \vspace{-10pt}
\end{figure*}

\begin{figure}[t]
\begin{center}
    \includegraphics[width=0.9\linewidth]{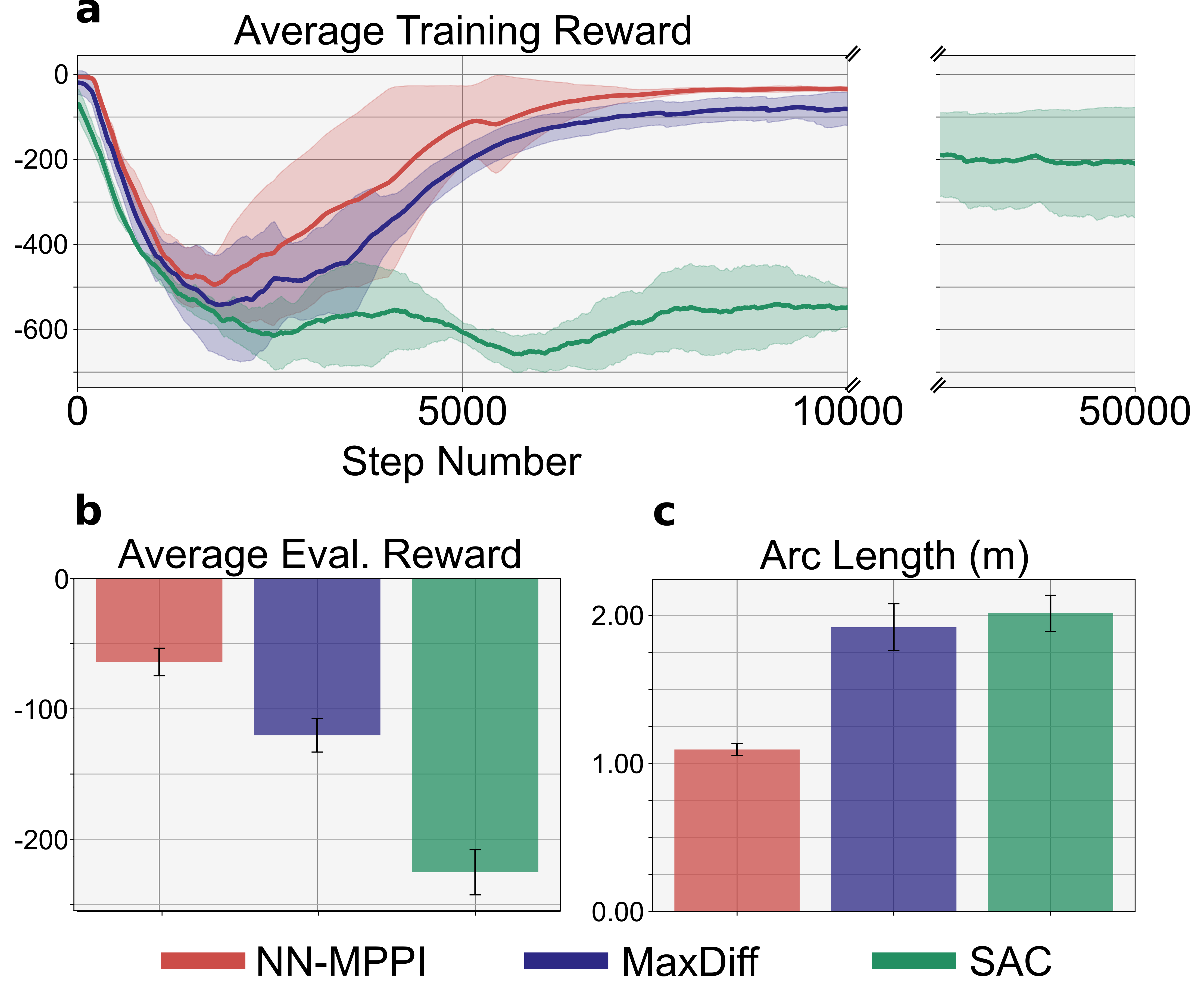}
    \caption{\textbf{Single-shot center balancing.} We report average training and evaluation performance for each algorithm across five seeds. \textbf{a:} Average position rewards during training. We trained SAC for 5x the duration in order to achieve the task. \textbf{b:} Average position rewards during evaluations. \textbf{c:} The average arc length of each evaluation trajectory---circuitous or oscillatory paths result in a high arc length.}
    \label{fig:single_metric_3}
    \vspace{-20pt}
\end{center}
\end{figure}

\subsection{Reinforcement Learning Environment}

In this subsection we describe the observation space, action space, and reward function of our RL environment. Both tasks use the same state and action variables. Our observation space $\textbf{o} \in \mathbb{R}^{9}$ is:
\begin{equation}
    \textbf{o} = [x, y, \dot{x}, \dot{y}, \phi, \theta, \psi, X, Y]
\end{equation}
In the observation space, $[x, y]$ and their derivatives correspond to the position and velocity of the puck relative to the platform center. $[\phi, \theta, \psi]$ represents the Euler angles of the platform relative to its base, and $[X, Y]$ corresponds to the balancing setpoint for the puck.

Our action space $\textbf{a} \in \mathbb{R}^{2}$ is continuous and consists of $\Phi$ and $\Theta$ corresponding to the roll and pitch state commands sent to the platform. To smooth platform actuation, we filter actions by applying a weighted average with the previous action command sent to the robot.

We use the same reward function for both balancing tasks. The reward function has three components: a position, velocity, and an action term, corresponding to $P$, $V$ and $A$ in (\ref{reward}). The scalar multipliers $a$, $b$, and $c$ are set to 250, 24, and 50 respectively.
\begin{equation}
    r = a P +  b V + c A
    \label{reward}
\end{equation}
\begin{equation}
    P = -1 \times ( 5(\Delta x) ) ^2 - ( 5(\Delta y) ) ^2
    \label{position reward}
\end{equation}
$\Delta x$ and $\Delta y$ correspond to the distances from the setpoint. Our velocity reward component is:
\begin{equation}
    V = -1 \times d^2 \times ([\dot{x}, \dot{y}] \boldsymbol{\cdot} [\Delta x, \Delta y]) ^ 2,
    \label{velocity reward}
\end{equation}
where $\boldsymbol{\cdot}$ represents the dot product, and $d$ represents the distance from the setpoint. We penalize velocities that are moving away from the setpoint and set this expression equal to zero when the dot product of (\ref{velocity reward}) is positive. The term for the action reward is:
\begin{equation}
    A = \frac{-1}{|\Delta x| + |\Delta y|} \times (\Phi^2 + \Theta^2),
    \label{action reward}
\end{equation}
where the units of $\Phi$ and $\Theta$ are normalized to $[-1, 1]$. The action reward provides a control cost to disincentivize large actions when close to the goal. Position, distance, and velocity terms are in meters and meters per second, normalized to approximately $[-2, 2]$ given the observed states of the platform-puck system.

\begin{figure*}[t]
    \centering
    \includegraphics[width=0.9\textwidth,keepaspectratio]{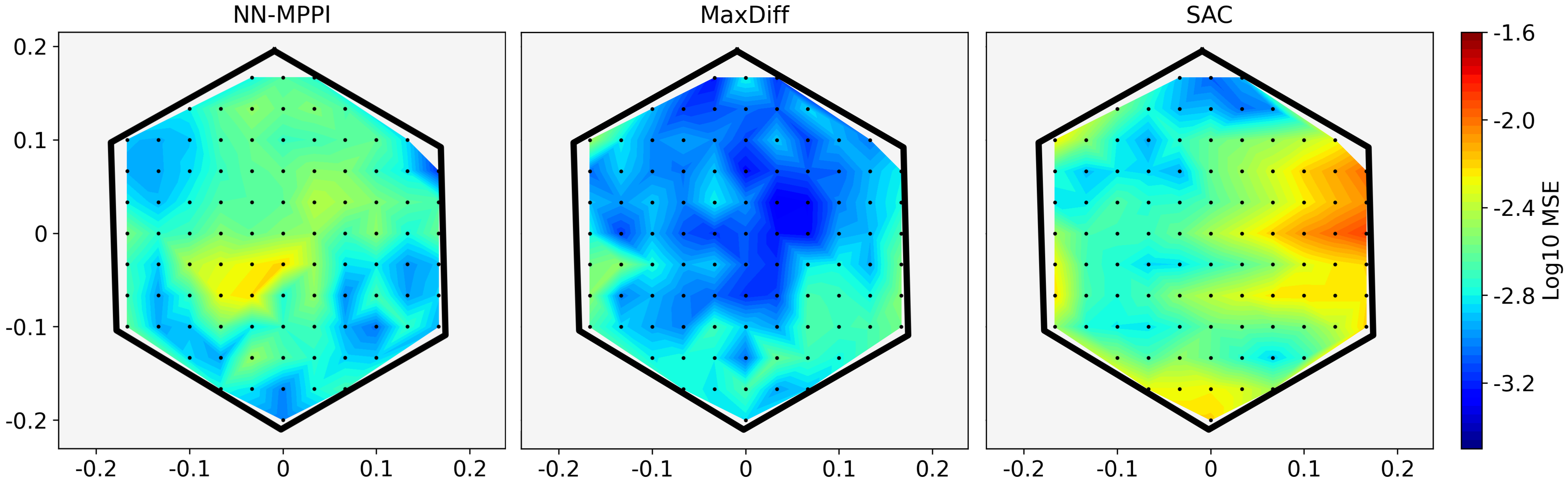}
    \caption{\textbf{Balancing at arbitrary points on the platform.} The scale for mean-squared-error (MSE) is logarithmic. MaxDiff has an average MSE that is 38\% better than NN-MPPI and 61\% better than SAC. We also found that MaxDiff was more consistent, with an MSE std. deviation 40\% smaller than NN-MPPI and 73\% smaller than SAC. The large error on the right half of the platform for SAC is primarily due to the failure of one seed to learn that side of the platform. }
    \label{fig:mock4}
    \vspace{-10pt}
\end{figure*}

\subsection{Reinforcement Learning Algorithms}

We evaluate three state-of-the-art reinforcement learning algorithms: Model-Predictive Path Integral Control (NN-MPPI), Maximum Diffusion (MaxDiff), and Soft Actor-Critic (SAC) \cite{haarnoja2018soft, williams2017information, Berrueta2024}. We list the relevant hyperparameters in Table \ref{mppi}.

\subsubsection{Model-Predictive Path Integral Control (NN-MPPI)}
NN-MPPI is a model-based reinforcement learning algorithm which models agent dynamics using a neural network. It is an information-theoretic derivation of MPPI---a sampling-based model-predictive control algorithm \cite{williams2017information} \cite{williams2016aggressive}.

\subsubsection{Maximum Diffusion (MaxDiff)}
Maximum Diffusion is a model-based framework for RL that utilizes the mechanics of diffusion to decorrelate agent experiences. It applies modified rewards to incentivize agent behavior that satisfies the statistics of a diffusion process \cite{Berrueta2024}. Our implementation and hyperparameters of MaxDiff are identical to that of NN-MPPI, with the addition of MaxDiff's temperature parameter $\alpha$. We follow an annealing schedule where $\alpha$ decays after remaining constant during the first half of training. For arbitrary-point balancing, the annealing coincides with the conclusion of the curriculum portion of the single-shot trial.

\subsubsection{Soft-Actor Critic (SAC)}

Soft Actor-Critic is a model-free, off-policy RL algorithm based on maximum entropy frameworks \cite{haarnoja2018soft}.

\textbf{\begin{table}[htbp]
\vspace{-20pt}
\begin{center}
\caption{Reinforcement Learning Hyperparameters}
\begin{tabular}{|c|l|l|l|}
\hline
\multicolumn{1}{|l|}{\textbf{Algorithm}} & \textbf{Hyperparameter} & \textbf{\begin{tabular}[c]{@{}l@{}}Center\\ Balancing\end{tabular}} & \textbf{\begin{tabular}[c]{@{}l@{}}Arbitrary\\  Balancing\end{tabular}} \\ \hline
\multirow{7}{*}{\textbf{All}} & Action Smoothing & 0.3 & 0.7 \\ \cline{2-4} 
 & Action Frequency & 10 Hz & 15 Hz \\ \cline{2-4} 
 & Learning Rate & 0.0003 & 0.0003 \\ \cline{2-4} 
 & Batch Size & 128 & 128 \\ \cline{2-4} 
 & Steps per Setpoint & N/A & 900 \\ \cline{2-4} 
 & Curriculum Length & N/A & 50,000 \\ \cline{2-4} 
 & Eval. Steps / Setpoint & 300 & 150 \\ \hline
\multirow{6}{*}{\textbf{\begin{tabular}[c]{@{}c@{}}NN-MPPI\\ MaxDiff\end{tabular}}}
 & Training Steps & 10,000 & 100,000 \\ \cline{2-4}
 & Planning Horizon & 20 & 20 \\ \cline{2-4} 
 & Samples & 4096 & 4096 \\ \cline{2-4} 
 & $\lambda$ & 0.1 & 0.1 \\ \cline{2-4}
 & Model Layers & 200 x 200 & 200 x 200 \\ \cline{2-4} 
 & Discount ($\gamma$) & 0.95 & 0.95 \\ \hline
\textbf{MaxDiff} & $\alpha$ & 0.1 & 0.1 \\ \hline
\multirow{5}{*}{\textbf{SAC}} & Training Steps & 50,000 & 100,000 \\ \cline{2-4} 
 & Reward Scale & 0.1 & 0.1 \\ \cline{2-4} 
 & Smoothing ($\tau$) & 0.005 & 0.005 \\ \cline{2-4} 
 & Policy Layers & 256 x 256 & 256 x 256 \\ \cline{2-4} 
 & Discount ($\gamma$) & 0.99 & 0.99 \\ \hline
\end{tabular}
\label{mppi}
\end{center}
\vspace{-15pt}
\end{table}}

\vspace{-5pt}
\section{RESULTS AND DISCUSSION}

We evaluate the performance of each RL algorithm in both center and arbitrary-point balancing tasks.

\subsection{Center Balancing}

We train five different seeds of each algorithm to balance in the center. The initial condition of the puck is approximately $[0, 0]$, and the platform is initialized with no tilt. We benchmark each algorithm by evaluating stabilization to the center from each corner of the platform. Each training session is 16min long---though policies frequently converge much sooner, see Fig. \ref{fig:single_metric_3}a---and each evaluation trial takes 30s per corner.

We benchmark the algorithms' performances in Figs. \ref{fig:center_eval} and \ref{fig:single_metric_3}---NN-MPPI and MaxDiff are consistently able to learn the task in under 15 min. SAC in this setting is less sample efficient---it is unable to accomplish any balancing in such a short timespan. We report SAC results from 50,000 training steps, where it stabilizes inconsistently. When SAC does learn to stabilize, the platform movements tend to be slower and the paths more direct. NN-MPPI and MaxDiff balance consistently, though oscillate about the setpoint. As compared to NN-MPPI, MaxDiff is more likely to oscillate at higher frequencies and contact the walls while initially moving towards the center.

As our implementations of NN-MPPI and MaxDiff are identical, we can directly ascribe performance differences to MaxDiff's modified objective function. For stabilizing at the platform center, we find that the additional exploration produced by MaxDiff is less helpful for simpler tasks with globally optimal states (i.e., motionless at the fixed setpoint), and are achievable by NN-MPPI. The NN-MPPI model may already explore nearby puck states due to the noise induced by the HSA vibrations and nonlinearities, making MaxDiff's exploration term less valuable.

\subsection{Arbitrary-Point Balancing}

For benchmarking arbitrary-point balancing, we train five seeds for each method, approximately 2 hours and 45 minutes each. Each model is evaluated by attempting to balance at 98 points evenly spaced across the platform surface. Evaluation for each point lasts 10s; the first three are omitted to allow for stabilization. Fig. \ref{fig:mock4} displays a heatmap of log-mean-squared error (MSE) when balancing across the platform for each algorithm---MaxDiff and NN-MPPI outperform SAC in arbitrary point balancing by 61\% and 38\% respectively. Model-free RL is generally less sample efficient than model-free approaches \cite{moerland2023model}---task success is therefore more difficult to achieve in the timeframes for which soft actuators are reliably functional. Although we did not observe any unintended breakages of HSAs during this study, sample efficiency is an important consideration for applying RL to other, less durable soft actuators. Maximum Diffusion outperforms NN-MPPI by 38\%---where its emphasis on exploration is beneficial for learning a more complex task and reward function that is dependent on the setpoint coordinates. Qualitatively, MaxDiff is able to consistently stabilize across the entire platform, whereas NN-MPPI is prone to missing small clusters of points and SAC frequently fails to balance on entire sections of the platform. Despite MaxDiff formally providing asymptotic guarantees in state-space exploration \cite{Berrueta2024}, we find that in practice a curriculum learning approach is necessary to balance at arbitrary coordinates. Without imposing additional structure on the RL, the puck frequently becomes stuck in a corner of the platform, especially when the initial sequence of setpoints is far from the platform center. Without a reset procedure for the environment, this can be catastrophic for single-shot learning, making the task incredibly challenging. With our curriculum approach and using MaxDiff, we observe consistent performance across random seeds and initializations. A numerical summary of our benchmarking is also shown in Table \ref{allresults}.

\begin{table}[hbtp]
\vspace{-5pt}
\begin{center}
\caption{Summary of reinforcement learning benchmarking}
\label{allresults}
\begin{tabular}{|c|ll|ll|}
\hline
\multicolumn{1}{|l|}{} & \multicolumn{2}{c|}{\textbf{Center Balancing}}           & \multicolumn{2}{c|}{\textbf{Arbitrary Balancing}}        \\ \cline{2-5} 
 & \multicolumn{1}{c|}{\textbf{Avg. Error}} & \multicolumn{1}{c|}{\textbf{Avg. Std.}} & \multicolumn{1}{c|}{\textbf{Avg. Error}} & \multicolumn{1}{c|}{\textbf{Avg. Std.}} \\ \hline
\textbf{NN-MPPI}       & \multicolumn{1}{l|}{\textbf{1.98 cm}} & \textbf{0.94 cm} & \multicolumn{1}{l|}{5.26 cm}          & 1.13 cm          \\ \hline
\textbf{MaxDiff}       & \multicolumn{1}{l|}{3.72 cm}          & 2.75 cm          & \multicolumn{1}{l|}{\textbf{4.37 cm}} & \textbf{0.91 cm} \\ \hline
\textbf{SAC}           & \multicolumn{1}{l|}{6.38 cm}          & 5.83 cm          & \multicolumn{1}{l|}{6.51 cm}          & 1.71 cm          \\ \hline
\end{tabular}
\end{center}
\vspace{-20pt}

\end{table}

\color{black}

\subsection{Control with Buckled and Broken Actuators}

\begin{figure}[htp]
    \centering
    \includegraphics[width=0.7\linewidth,keepaspectratio]{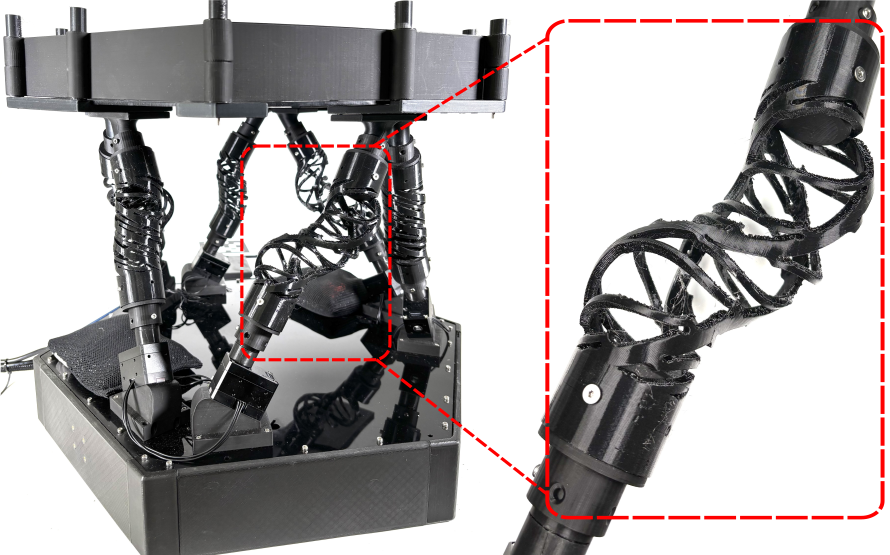}
    \caption{{\textbf{Learning with buckled actuators.} Buckling of HSAs occurs when a subset of the actuators are extended a significant range past their neighbors. When buckled, HSAs have reduced stiffness and approach an effective singularity---further rotation has little to no effect on the platform state---\color{black}until they are retracted to remove the buckling-induced hysteresis. Reliable single-shot RL outcomes can still be achieved despite triggering a buckling condition midway through training.
    }}
    \label{fig:buckled_platform}
    \vspace{-10pt}
\end{figure}

To demonstrate how RL can exploit larger configuration spaces in soft robotics, we buckle three of the platform's six HSAs by extending their neutral points by 240 degrees while training to balance the puck at the center, shown in Fig. \ref{fig:buckled_platform}. While the platform is still formally controllable due to its over-actuation for this positioning task, the dynamics of the buckled actuators are nonlinear and stochastic based on loading conditions and the path-dependent, hysteretic effects of the transmissions.

\begin{figure}[htp]
    \centering
    \vspace{10pt}    \includegraphics[width=0.7\linewidth,keepaspectratio]{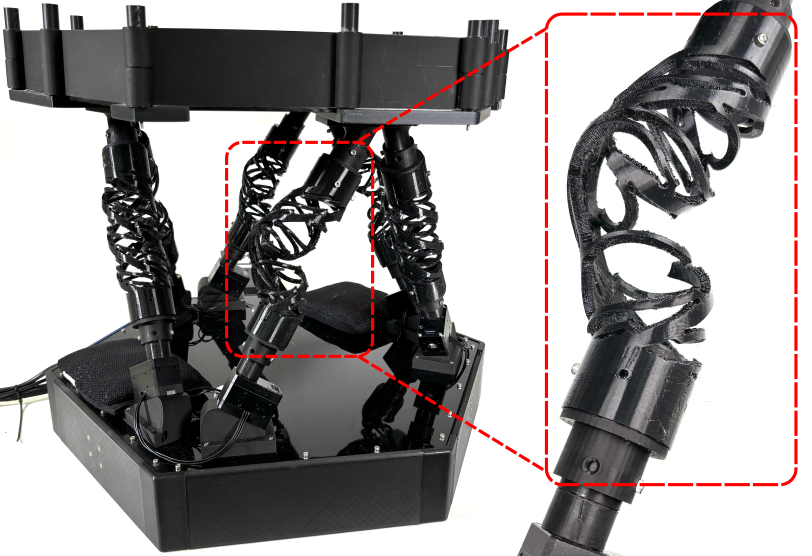}
    \caption{{\textbf{Learning with broken actuators.} Handed Shearing Auxetics are able to suffer significant breakages of their living hinges and still retain partial functionality. Despite cutting the major helices of half of the HSAs during training, single-shot RL is able to achieve near-identical learning outcomes when compared against unbroken actuators.}}
    \label{fig:broken_platform}
    \vspace{-10pt}
\end{figure}

Furthermore, HSAs are especially useful as soft transmissions for their ability to incur significant mechanical damage and still retain partial functionality. This is advantageous in contrast to pneumatics, fluidics, and some cable-driven actuators that may have a single point of failure. However, when broken, HSAs exhibit reduced and asymmetrical bending stiffness, and reduced axial stiffness. We show how RL can accommodate wear and adversarial damage as shown in Fig. \ref{fig:broken_platform} by breaking the major helices on three HSAs during a training episode.

\color{black}

For each of these comparisons, we train and evaluate five seeds  of MaxDiff policies for center balancing, performing each dynamics change halfway through the episode and resuming training without any reset behavior\color{black}. We use MaxDiff for this demonstration as \cite{Berrueta2024} shows how single-shot policies are able to generalize across embodiments or shifts in dynamical properties, with benchmarking showing an improvement over NN-MPPI.

\begin{figure}[hbtp]
    
    \centering
    \includegraphics[width=0.9\linewidth,keepaspectratio]{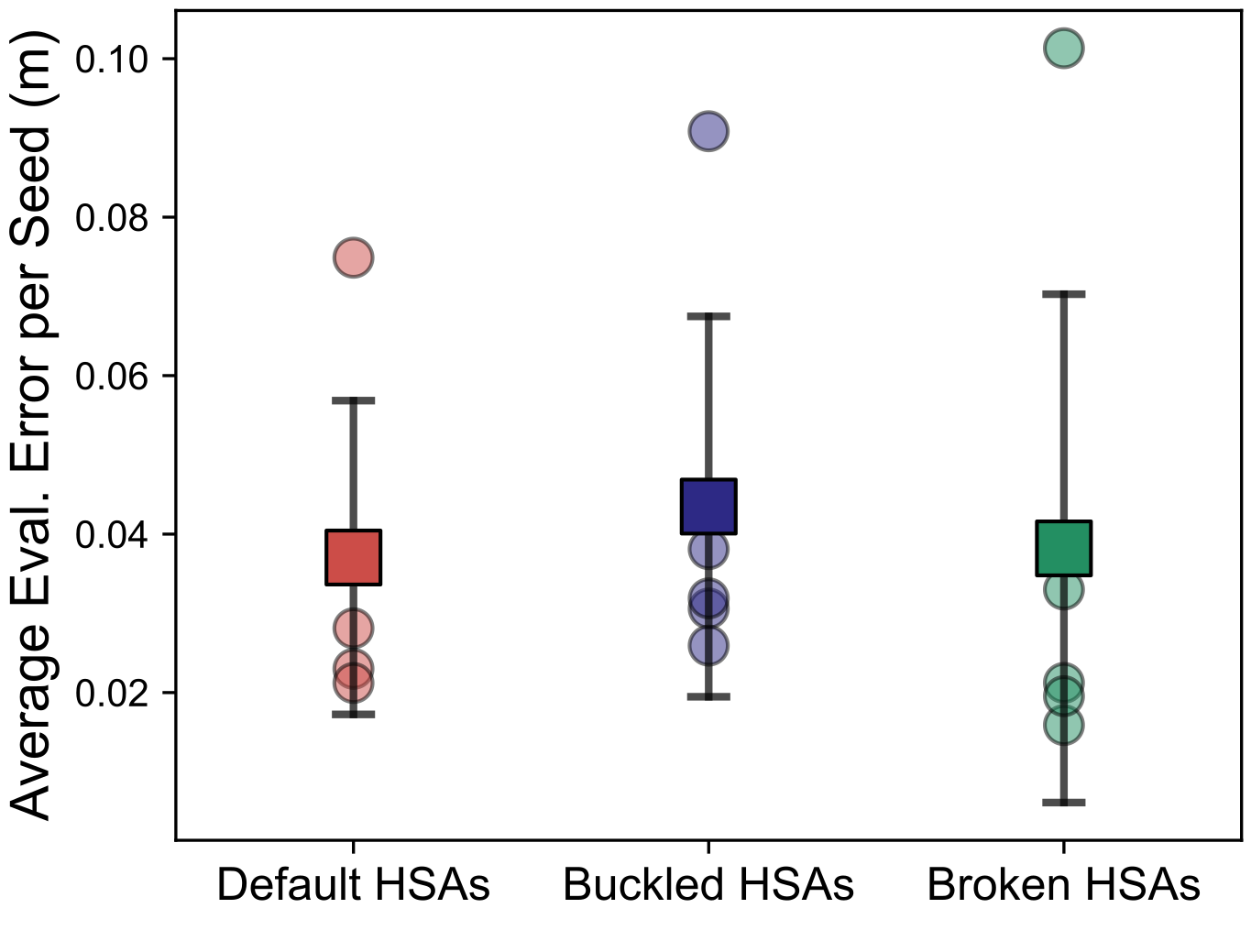}
    \vspace{-10pt}
    \caption{{\textbf{Modifying actuators during training.} At the midpoint of each training episode, half of the HSAs are either buckled or broken. In a single episode, RL is able to overcome these changes and has near-identical evaluation performance to the default, unmodified HSAs. Each seed consists of six evaluation trials and distances are calculated using the last 10 seconds of stabilization.}}
    \label{buckle_comparison}
    \vspace{-10pt}
\end{figure}

\color{black}
In both cases of shifting dynamics---one in an out-of-distribution portion of the configuration space, and the other with adversarial damage, RL is able to achieve results consistent with the actuators fixed to their default states as shown in Fig. \ref{buckle_comparison}, despite teleoperating the robot becoming qualitatively more challenging. Further investigation into both explicitly sensing and adapting to damage and utilizing larger volumes of the platform's configuration space remains the subject of future work.

\color{black}

\section{CONCLUSIONS}

We present the first demonstration of single-shot reinforcement learning in a soft robot, relying entirely on hardware and without any simulation. By using curriculum learning, we show that controllers for dynamic balancing tasks can be learned reliably in a single episode. We demonstrate single-shot learning under environmental stress by buckling and breaking actuators with a bolt cutter during training---see associated multimedia---and achieving near-identical learning outcomes. We thoroughly benchmark the model-based and model-free RL approaches, providing considerations for real-world learning in soft robots whose actuator properties evolve over time and whose workspaces are traditionally limited to avoid nonlinearity. Future work will focus on sample-efficient approaches that explicitly incorporate the introduction of modulating stiffness and hysteresis to control greater volumes of the robot's configuration space for more diverse tasks and for applications that rely upon large deformations and environmental contact. Using RL to reliably learn policies that generalize to new and evolving environments in single episode learning on hardware will enable soft robots to complete many practical tasks in the real world that would otherwise be computationally infeasible.
\color{black}






\bibliographystyle{ieeetr}

\bibliography{export}

\end{document}